%% file: main.tex
\documentclass{article}
\usepackage[utf8]{inputenc}
\usepackage{graphicx} %
\usepackage{natbib}
\usepackage{bbm}
\usepackage{amsmath}
\usepackage{amssymb}
\usepackage{graphicx}
\usepackage{lipsum}
\usepackage{booktabs}
\usepackage{siunitx}
\usepackage{multirow}
\usepackage{adjustbox}
\usepackage{parskip}

\PassOptionsToPackage{hyphens}{url}\usepackage{hyperref}

\title{Traveling Words: A Geometric Interpretation of Transformers}
\author{Raul Molina \\ \href{mailto:mail@rsmolina.com}{mail@rsmolina.com}}
\date{September 2023}

\begin{document}

\maketitle

\input{00abstract}
\input{01intro}
\input{03body}

\input{04exps}

\input{05conclusion}

\bibliography{references.bib}

\newpage

\input{06appendix}

\newpage

\end{document}

%% file: 00abstract.tex
\section{Abstract}

Transformers have significantly advanced the field of natural language processing, but comprehending their internal mechanisms remains a challenge.
In this paper, we introduce a novel geometric perspective that elucidates the inner mechanisms of transformer operations.
Our primary contribution is illustrating how layer normalization confines the latent features to a hyper-sphere, subsequently enabling attention to mold the semantic representation of words on this surface.
This geometric viewpoint seamlessly connects established properties such as iterative refinement and contextual embeddings.
We validate our insights by probing a pre-trained 124M parameter GPT-2 model.
Our findings reveal clear query-key attention patterns in early layers and build upon prior observations regarding the subject-specific nature of attention heads at deeper layers.
Harnessing these geometric insights, we present an intuitive understanding of transformers, depicting them as processes that model the trajectory of word particles along the hyper-sphere.

%% file: 01intro.tex
\section{Introduction}

The transformer architecture \citep{vaswani2017attention} has sparked a significant shift in Artificial Intelligence (AI).
It is the central component behind some of the most advanced conversational AI systems \citep{brown2020language, thoppilan2022lamda, bai2022constitutional}, and has been established as state-of-the-art for Natural Language Processing (NLP), Computer Vision (CV) and Robotics applications, and many others \citep{OpenAI2023, Google2023, chen2023symbolic, zong2022detrs, driess2023palm}.

Recent work on the interpretability of the transformer architecture has focused on analyzing weights in relation to the word embedding space used in its input and output layers \cite{dar2022analyzing, elhage2021mathematical, geva2022transformer, brody2023expressivity, windsor2022layer, beren2022svd}.
\citet{elhage2021mathematical} introduces ``Transformer Circuits", a theoretical framework that decomposes the transformer computation into two main components: a residual stream that carries information from input to output layers and attention/feed-forward updates that modify the information flowing in the residual stream.
A key development from their work is the grouping of the $W_Q$$W_K^T$ and $W_O$$W_V^T$ matrices from the attention mechanism, representing low-rank approximations of virtual matrices $W_{QK}$ and $W_{OV}$, respectively. These virtual matrices define interactions between different words in the input sequence $X$ within a canonical feature space $E$ given by the word embedding matrix $W_E$.
The resulting values from these interactions are used to update the information carried throughout the residual stream. 
\citet{geva2022transformer} further decompose the operations within the Transformer, demonstrating that the updates from the feed-forward module can be decomposed into a linear combination of sub-updates given by the weight matrix of the feed-forward module's second layer $W_2$.
The matrix $W_2$ also interacts within the canonical space $E$ and allows the authors to measure the impact of each sub-update on the model's final prediction using the matrix $W_E$ as a probe.
\citet{dar2022analyzing} incorporate these ideas to show that it is not only possible to interpret the outcomes of each Transformer operation in relation to the canonical space $E$ but also the weights themselves, enabling them to do zero-shot model stitching by ``translating" between the canonical spaces of different language models.
Finally, \citet{beren2022svd} note that analysis on the singular vectors of the $W_{OV}$ matrices provides better practical results when compared to analysis of its row and column weights.

A complimentary perspective to the line of work on Transformer Circuits comes from the geometric interpretation of  layer normalization \citep{ba2016layer} by \citet{brody2023expressivity}.
The authors prove that layer normalization is equivalent to projecting features onto the hyperplane defined by the $\overrightarrow{\mathbbm{1}}$ vector and then scaling the projection by $\sqrt{d}$. 
They show that these properties are crucial for the attention mechanism to either attend to all keys equally or to avoid the problem of having ``unselectable" keys (relevant keys within the convex hull of a set of non-relevant keys).
\citet{windsor2022layer} provides further evidence of the representational power of layer normalization, visualizing the highly non-linear behavior that arises from this operation.
The authors demonstrate that, when used as an activation function within a neural network, layer normalization is capable of solving complex classification tasks.

\begin{figure}[htbp]
    \centering
    \includegraphics[width=0.5 \linewidth]{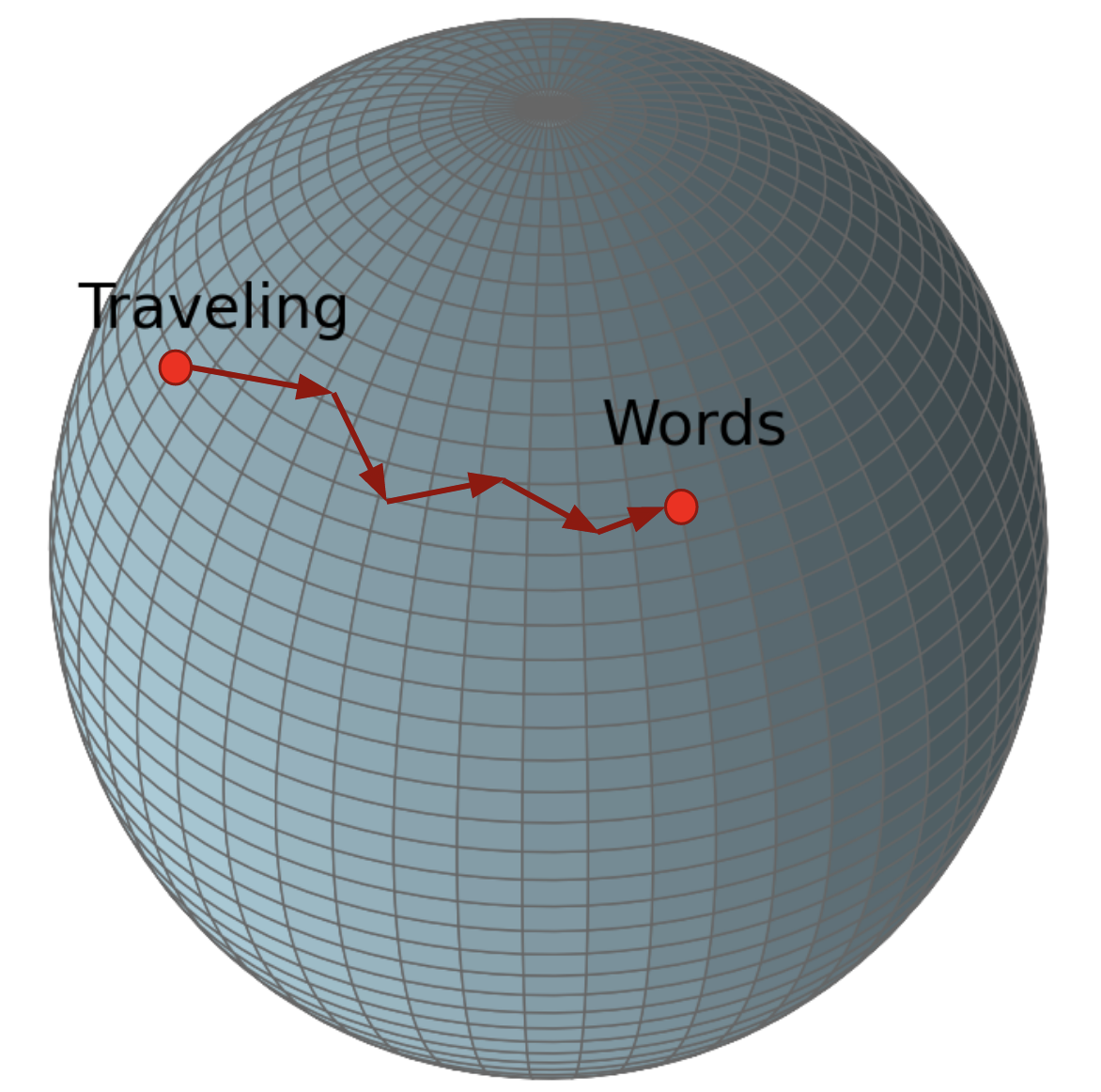}
    \caption{Overview of the proposed geometric interpretation of Transformers. In it, the phrase ``Traveling Words" is to be completed by a given transformer model. The input token ``Traveling " is embedded as a word particle using an embedding matrix $W_E$ and projected onto a hyper-sphere  using layer normalization. Each subsequent layer in the transformer determines the path that the particle will follow along the surface of the hyper-sphere, culminating on the region closest to the next token: ``Words".}
    \label{fig:traveling}
\end{figure}

In this work, we connect these ideas under a single interpretation: word particles traveling around the surface of a hyper-sphere, completing a journey that goes from a previous word to the next and transforming their meaning throughout this process.
An illustrated summary of this interpretation is given in \autoref{fig:traveling}.

%% file: 03body.tex
\section{Transformers as a Composition of Geometric Primitives}

In this section, we analyze each of the transformer's components from a geometric perspective, leveraging the interpretation of one component to analyze the next.
We begin with the layer normalization function, for which we demonstrate that it constrains $d$-dimensional input features to lie within the surface of a $(d-1)$ dimensional hyper-sphere.
Then we consider the role of the $W_{QK}$ matrix in terms of geometric transformations on said hyper-sphere, and $W_{VO}$ as a key-value mapping from the hyper-sphere back to $\mathbb{R}^d$.
Additionally, we review the key-value interpretation of the feed-forward module proposed by \citet{geva2021transformer}.
Finally, we discuss the role of the embedding matrix $W_E$ on the transformer's output probabilities.

\subsection{Layer Normalization}\label{sec:layer_norm}

\begin{figure}[h]
    \centering
    \includegraphics[width=\linewidth]{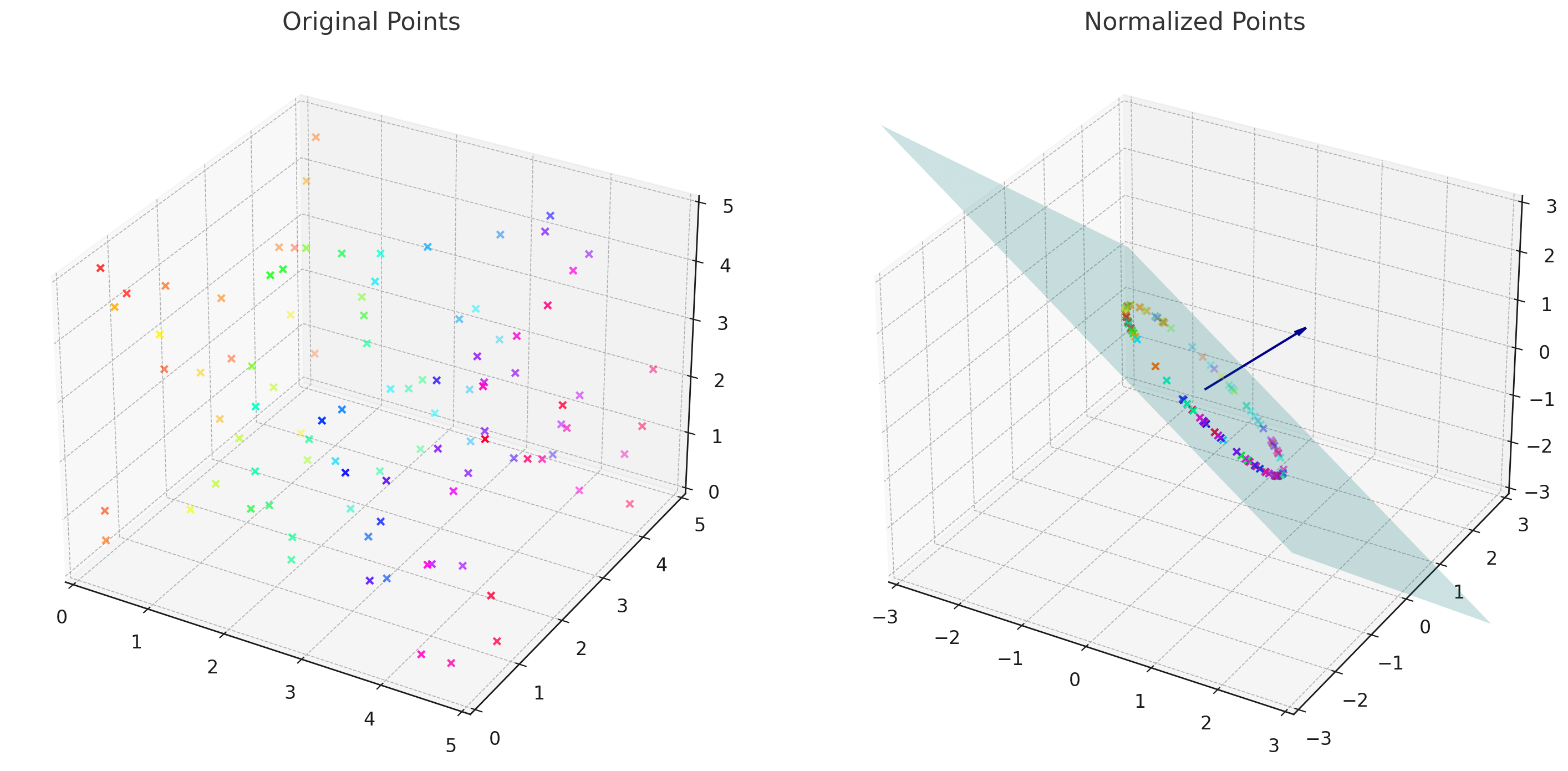}
    \caption{
    Layer normalization visualized on 3D data.
    Left: Original feature space (from randomly sampled data), with each data point color-coded according to its position in space.
    Right: Feature space after layer normalization, note that all data points lie within the plane perpendicular to the $\overrightarrow{\mathbbm{1}}$ vector.
    }
    \label{fig:layernorm}
\end{figure}

In its original formulation \citep{ba2016layer}, layer normalization is introduced in terms of the mean $\mu$ and standard deviation $\sigma$ of an input feature vector $X \in \mathbb{R}^d$:
\begin{equation}
    \label{eq:ln_stats}
    \text{LayerNorm}(X) = \frac{X - \mu}{\sigma}
\end{equation}
Where both the mean and standard deviation are taken along the feature dimension $d$ such that:
\begin{align*}
    \mu &= \frac{1}{d} \sum_i^d x_i \\
    \sigma &= \sqrt{\frac{1}{d} \sum_i^d (x_i - \mu)^2} \\
\end{align*}
\\
\citet{brody2023expressivity} note that the numerator in \autoref{eq:ln_stats} is itself an operation between $X$ and the vector $\boldsymbol{\mu}$ defined as:
\begin{equation*}
    \boldsymbol{\mu} = [ \mu, \mu, \dots, \mu ] \in \mathbb{R}^d
\end{equation*}
\\
The resulting vector $(X - \boldsymbol{\mu})$ is shown to be orthogonal to the $\overrightarrow{\mathbbm{1}}$ vector, and thus layer normalization can be interpreted as a projection of $X$ onto the hyperplane defined by the normal vector $\overrightarrow{\mathbbm{1}}$.
\citet{brody2023expressivity} also show that the division by $\sigma$ acts as a scaling factor that modifies the norm of $(X - \boldsymbol{\mu})$ to be $\sqrt{d}$:
\begin{align}
\label{eq:sigma}
\begin{split}
    \sigma &= \sqrt{\frac{1}{d}\sum_i^d (x_i - \mu)^2}\\
    &= \frac{1}{\sqrt{d}} \sqrt{\sum_i^d(x_i - \mu)^2}\\
    &= \frac{1}{\sqrt{d}} ||X-\boldsymbol{\mu}||_2
\end{split}
\end{align}
\\
We note that, if we consider unit-norm vector $\frac{1}{\sqrt{d}} \overrightarrow{\mathbbm{1}}$ instead of  $\overrightarrow{\mathbbm{1}}$, it can be shown that $\boldsymbol{\mu}$ is the projection of $X$ onto $\frac{1}{\sqrt{d}} \overrightarrow{\mathbbm{1}}$ (explaining why $X - \boldsymbol{\mu}$ is orthogonal to $\overrightarrow{\mathbbm{1}}$):
\begin{align}
\begin{split}
    \text{proj}(X, \frac{1}{\sqrt{d}} \overrightarrow{\mathbbm{1}}) &= \frac{1}{||\frac{1}{\sqrt{d}} \overrightarrow{\mathbbm{1}}||_2} \Bigg( \mathbf{x} \cdot \frac{ 1}{\sqrt{d}} \overrightarrow{\mathbbm{1}} \Bigg) \frac{1}{\sqrt{d}} \overrightarrow{\mathbbm{1}}\\
    &= \Big( \frac{\mathbf{x} \cdot \overrightarrow{\mathbbm{1}}}{\sqrt{d}} \Big) \frac{1}{\sqrt{d}} \overrightarrow{\mathbbm{1}}\\
    &= \Big( \frac{\mathbf{x} \cdot \overrightarrow{\mathbbm{1}}}{d} \Big) \overrightarrow{\mathbbm{1}}\\
    &= \Big(\frac{1}{d} \sum_i^d x_i \Big) \overrightarrow{\mathbbm{1}}\\
    &= \mu \overrightarrow{\mathbbm{1}}\\
    &= \boldsymbol{\mu}
\end{split}
\end{align}
\\
From this result, it is straightforward to calculate the projection of $X$ onto the hyperplane $\mathcal{H}$ defined by $\frac{1}{\sqrt{d}} \overrightarrow{\mathbbm{1}}$:

\begin{align}
\label{eq:proj_h}
\begin{split}
    \text{proj}_{\mathcal{H}}(X) &= X - \text{proj}(X, \frac{1}{\sqrt{d}} \overrightarrow{\mathbbm{1}})\\
    &= X - \boldsymbol{\mu}
\end{split}
\end{align}
\\
Finally, we can use the results from \autoref{eq:sigma} and \autoref{eq:proj_h} to reformulate layer normalization in geometric terms:
\begin{align}
\label{eq:ln_geom}
\begin{split}
    \text{LayerNorm}(X) &= \frac{X - \mu}{\sigma}\\
    &= \frac{\text{proj}_{\mathcal{H}}(X)}{ \frac{1}{\sqrt{d}}||\text{proj}_{\mathcal{H}}(X)||_2}\\
    &=\sqrt{d} \; \frac{\text{proj}_{\mathcal{H}}(X)}{||\text{proj}_{\mathcal{H}}(X)||_2}
\end{split}
\end{align}
\\
Intuitively, layer normalization projects a vector $X \in \mathbb{R}^d$ to the hyperplane $\mathcal{H}$ perpendicular to $\overrightarrow{\mathbbm{1}} \in \mathbb{R}^d$, and normalizes the projection such that it lies on the surface of a $d-1$ dimensional hyper-sphere of radius $\sqrt{d}$.
A visualization of this process for $d=3$ is shown in \autoref{fig:layernorm}.
In practice, layer normalization includes two additional parameters: a scaling factor $\gamma$ and a bias term $\beta$.
The parameter $\gamma$ scales each coordinate axis of $\mathbb{R}^d$ independently, transforming the hyper-sphere into a hyper-ellipsoid, and the bias term $\beta$ shifts the center of said ellipsoid away from the origin.
A 2D representation of the entire process is shown in \autoref{fig:w_qk}.

\citet{xiong2020layer} show that layer normalization should be applied within each block before the attention and feed-forward module updates and as a final step before prediction.
Doing so scales down the gradient of the feed-forward module's weight parameters and keeps the magnitude of the hidden states bounded with respect to the depth of a given layer, which improves stability during training and removes the need for a warm-up stage.

\subsection{Multi-Head Self-Attention}\label{sec:multihead_attention}

In the previous section, we showed how the layer normalization approach given by \citet{xiong2020layer} enforces data within each layer to be constrained on the surface of a hyper-sphere, potentially unique to each layer. However, thanks to the residual nature of transformers, all intermediate layer representations share the same vector space and thus are essentially projecting features onto the same hyper-sphere $\mathcal{H}_S$. Furthermore, given that layer normalization is applied before the classification softmax, the model maximizes the dot-product similarity between a subset of points within $\mathcal{H}_S$ and the word vectors in the embedding matrix $W_E \in \mathbb{R}^{|V| \times d}$ (where $|V|$ denotes the size of the vocabulary), establishing the meaning of points in $\mathcal{H}_S$ in relation to the words associated with $W_E$.
To understand how the geometric intuition behind $\mathcal{H}_S$ allows for interpretability, we will first revisit the self-attention module in transformers \citep{vaswani2017attention}.

For a given input sequence $X \in \mathbb{R}^{s \times d}$ of length $s$, the self-attention mechanism is defined as follows:
\begin{equation}
    \text{SelfAttention}(X, W_Q, W_K, W_V) = \text{softmax}\Big(\frac{Q K^T}{\sqrt{d}}\Big) V
\end{equation}
where
\begin{align}
\label{eq:qkv_matrices}
\begin{split}
    Q &= X W_Q\\
    K &= X W_K\\
    V &= X W_V
\end{split}
\end{align}
\\
Such that $W_Q \in \mathbbm{R}^{d \times k}$, $W_K \in \mathbbm{R}^{d \times k}$ and $W_V \in \mathbbm{R}^{d \times v}$ are projection matrices from the original model dimension $d$ to intermediate dimension $k$ and value dimension $v$, respectively.
For multi-head attention, multiple projection matrices $W_Q^i$, $W_K^i$, $W_V^i$ are considered, one for each head $i \in [1, \dots, h]$ (with $h$ being the number of heads).
In this case, the value dimension $v$ is commonly set equal to $k$ and an extra projection matrix $W_O \in \mathbb{R}^{hk \times d}$ is introduced to combine information from all heads as follows \citep{vaswani2017attention}:
\begin{align}
\label{eq:multihead_original}
\begin{split}
    \text{MultiHead}(X) &= \text{Concat}([\text{head}_1, \dots, \text{head}_h]) W_O\\
    \text{where} \; \text{head}_i &= \text{SelfAttention}(X, W_Q^i, W_K^i, W_V^i)
\end{split}
\end{align}
\\
Given that the concatenation happens along the row dimension of each head, it is possible to re-write multi-head self-attention as follows:
\begin{align}
\label{eq:multihead_sum}
\begin{split}
    \text{MultiHead(X)} &= \sum_i^h \text{SelfAttention}(X, W_Q^i, W_K^i, W_V^i) W_O^i\\
    \text{where} \; W_O &= \text{Concat}[W_O^1, \dots, W_O^h]
\end{split}
\end{align}
\\
Such that each $W_O^i \in \mathbb{R}^{k \times d}$ denotes an element of the partition of matrix $W_O$ alongside the row dimension.
Combining \autoref{eq:qkv_matrices} and \autoref{eq:multihead_sum} we obtain a single formula for multi-head self-attention:
\begin{align}
\label{eq:multihead_final}
\begin{split}
    \text{MultiHead(X)} &= \sum_i^h \text{softmax}\Bigg(\frac{X W_Q^i {W_K^i}^T X^T}{\sqrt{d}}\Bigg) X W_V^i W_O^i \\
    &= \sum_i^h \text{softmax}\Bigg(\frac{X W^i_{QK} X^T}{\sqrt{d}}\Bigg) X W^i_{VO}
\end{split}
\end{align}
Where $W_{QK}^i \in \mathbb{R}^{d \times d}$ and $W_{VO}^i \in \mathbb{R}^{d \times d}$ are low-rank virtual matrices obtained by grouping $W_Q^i W_K^{iT}$ and $W_V^i W_O^i$ respectively \citep{elhage2021mathematical, dar2022analyzing}.

\subsection{The \texorpdfstring{$W_{QK}$}{QK} Matrix}\label{sec:w_qk_matrix}

\begin{figure}[htbp]
    \centering
    \includegraphics[width=\linewidth]{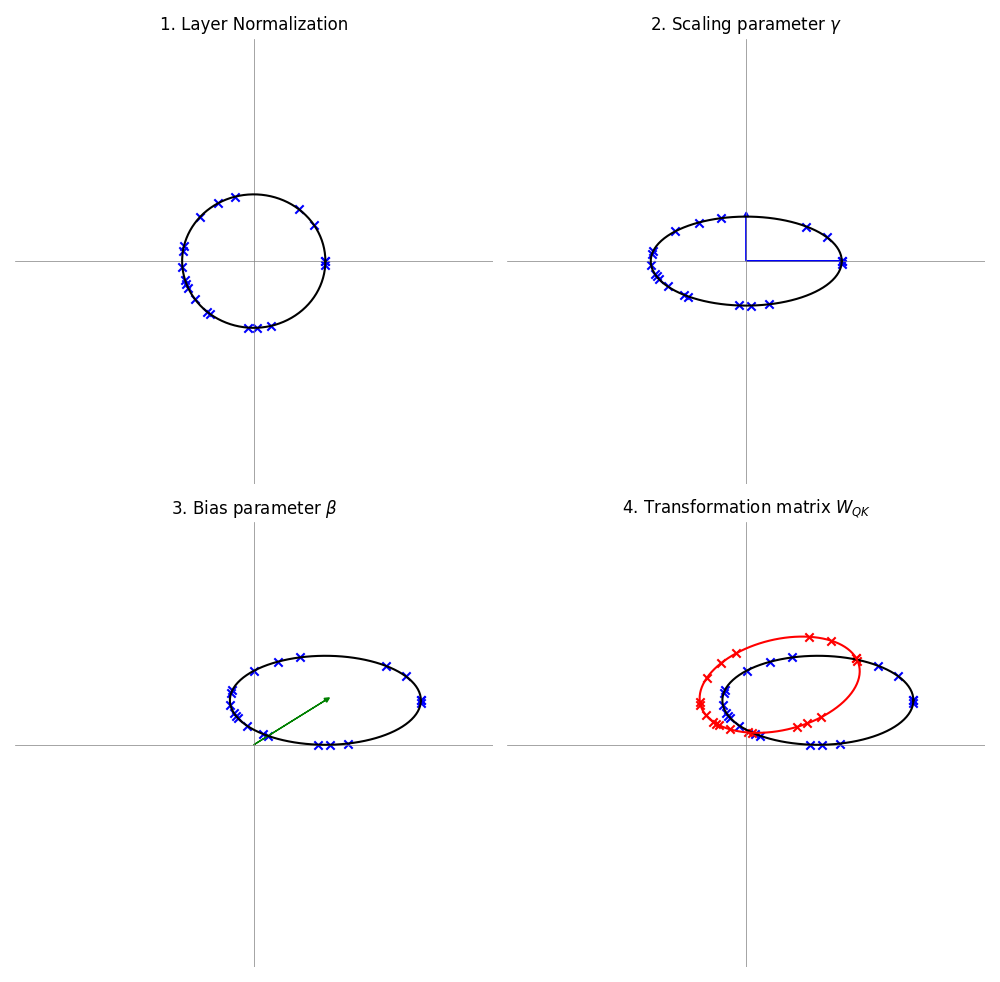}
    \caption{
        Visualization of the self-attention process for a single head. \textbf{Top Left}: Layer normalization projects the input features on the surface of the hyper-sphere $\mathcal{H}_S$. \textbf{Top Right}: A scaling parameter $\gamma$ is commonly applied after normalization; it transforms $\mathcal{H}_S$ into an ($d-1$)-dimensional ellipsoid. \textbf{Bottom Left}: A bias term $\beta$ is also applied after normalization; it displaces the ellipsoid away from the origin. \textbf{Bottom Right}: The input features are transformed to a query representation (in red) using the $W_{QK}$ matrix and compared against their previous representation to obtain the self-attention scores.
    }
    \label{fig:w_qk}
\end{figure}

For any given head, the matrix $W^i_{QK}$ is commonly interpreted as a bi-linear form $f: \mathbb{R}^d \times \mathbb{R}^d \rightarrow \mathbb{R}$ that represents the relevance between keys and queries.
However, it is also possible to consider $W^i_{QK}$ as a linear transformation that maps inputs to a query representation $X^i_q \in \mathbb{R}^{s \times d}$ (similar to that considered in \cite{brody2023expressivity}):
\begin{equation}
    XW^i_{QK} = X^i_{q}
\end{equation}
\\
With the head's attention score matrix $A^i \in [0, 1]^{s \times s}$, for a given sequence length $s$, obtained as:
\begin{equation}
\label{eq:attention_matrix}
    A^i = \text{softmax}\Big(\frac{X^i_{q} X^T}{\sqrt{d}}\Big)
\end{equation}
\\
Alternatively, its transpose can be considered as a transformation that maps inputs to a key representation:
\begin{equation}
    X{W^i_{QK}}^T = X^i_{k}
\end{equation}
With the attention score matrix as follows:
\begin{equation}
    A^i = \text{softmax}\Big(\frac{X (X^i_{k})^T}{\sqrt{d}}\Big)
\end{equation}
This process is illustrated for normalized inputs in the bottom-right section of \autoref{fig:w_qk}.
Essentially, the role of the $W_{QK}$ matrix and the layer normalization parameters is to find a transformation over $\mathcal{H}_S$ such that, when superimposed on itself, brings related terms closer together and keeps unrelated terms apart.

It is important to mention that for $k < d$, the matrix $W^i_{QK}$ cannot be inverted, as it won't have a full rank.
This implies (by the rank-nullity theorem) that for each head, there must be a set of $d-k$ query vectors $Q^i_{null} \subset \mathbb{R}^d$ that map to the zero vector and, as a consequence, attend to all keys equally.
Conversely, there must also exist a set of $d-k$ keys $K^i_{null} \subset \mathbb{R}^d$ that are attended to by all queries equally, with an attention score of zero.

\textbf{Note on bias terms:} In case the projection given by \autoref{eq:qkv_matrices} contains bias terms $\beta_q$, $\beta_k \in \mathbb{R}^{1 \times k}$, the attention score matrix from \autoref{eq:attention_matrix} is calculated as follows:
\begin{equation}
    \label{eq:attention_matrix_with_bias}
    A^i = \text{softmax}\Big( \frac{X^i_{q} X^T + X W^i_Q \beta_k^T + \beta_q {W^i_k}^T X^T + \beta_q \beta_k^T}{\sqrt{d}} \Big)
\end{equation}
\\
In the bias formulation, three new terms are introduced.
First, $W^i_Q \beta_k^T \in \mathbb{R}^{d \times 1}$, which can be thought of as a reference vector for queries, such that queries similar to it get higher attention scores.
Given that the same ``bias score" will be broadcasted along all the different keys of the same query, the network will ignore this term due to the shift-invariance of the softmax function.
More interesting is the second term $\beta_q {W^i_k}^T \in \mathbb{R}^{1 \times d}$,  which acts as a reference for keys.
Given that its bias score is broadcasted along queries, it will result in higher attention scores (in all queries) for keys similar to the reference.
Finally, the term $\beta_q \beta_k^T \in \mathbb{R}$ acts as a global bias and, similar to $W^i_Q \beta_k^T$, will be ignored by the network.

\subsection{The \texorpdfstring{$W_{VO}$}{VO} Matrix and the Residual Stream}\label{sec:w_vo_matrix}

\begin{figure}[htbp]
    \centering
    \includegraphics[width=0.8\linewidth]{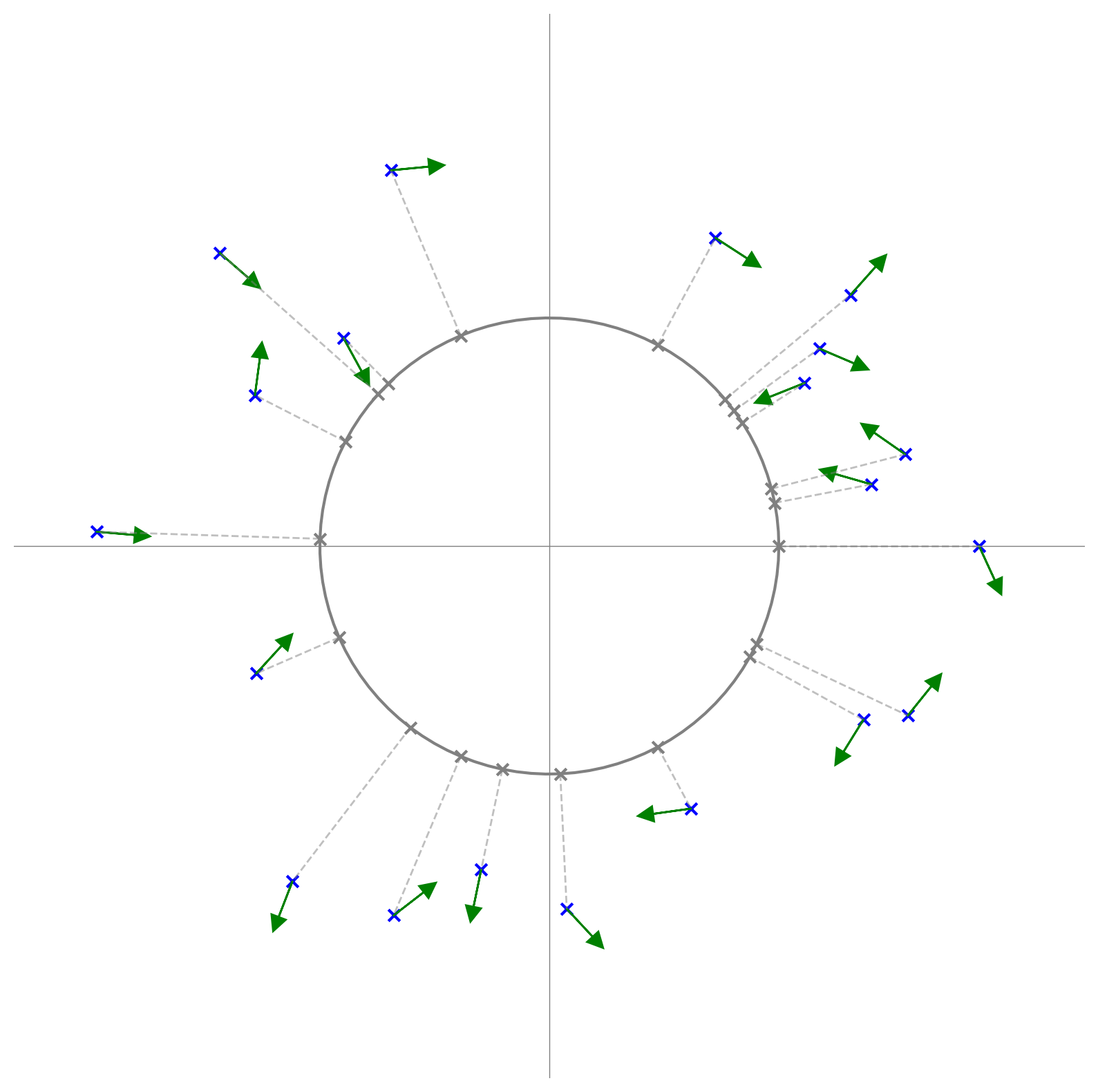}
    \caption{
        Visualization of the residual update for a single attention head. For each normalized data point (in gray), there is a corresponding un-normalized data point in the residual stream (in blue). Data points in the residual stream are updated according to a given direction (in green) calculated from the self-attention scores and the update matrix $W_{VO}$. 
    }
    \label{fig:w_vo}
\end{figure}

To understand the role of the $W_{VO}$ matrix within the transformer, we now consider the update step after the multi-head attention layer:
\begin{equation}
    X_{l + 1} = X_l + \text{MultiHead}(\text{LayerNorm}(X))
\end{equation}
\\
Note that by plugging in \autoref{eq:multihead_final} and \autoref{eq:attention_matrix}, the layer update can be re-written as:
\begin{equation}
\label{eq:layer_update}
    X_{l + 1} = X_l + \sum_i^h A^i X^i_{value}
\end{equation}
\\
where
\begin{equation}
\label{eq:x_value}
    X^i_{value} = \text{LayerNorm}(X_l) W_{VO}^i
\end{equation}
\\
It can be seen that the multi-head attention mechanism consists of the sum of $h$ individual updates, each one given by one of the attention heads.
Within each head, all words in the sequence propose an update $X^i_{value}$, and these are aggregated according to their attention scores $A^i$.
In \autoref{eq:x_value}, the matrix $W^i_{OV}$ acts as a map that takes the normalized inputs in $\mathcal{H}_S$ (adjusting for scale and bias) and outputs a set of updates in the same space as $W_E$, this process is visualised in \autoref{fig:w_vo}.
Furthermore, we propose that the $W^i_{VO}$ matrix is better understood as a second key-value store \citep{sukhbaatar2015end, geva2021transformer} within the attention layer.
To see why, consider its Singular Value Decomposition (SVD) \citep{beren2022svd}:
\begin{equation}
    \label{eq:w_vo_svd}
    W^i_{VO} = U \Sigma V^T
\end{equation}
\\
By substituting in \autoref{eq:x_value}, we obtain: 
\begin{equation}
    X^i_{value} = (Q_{VO} {K^i_{OV}}^T) V^i_{OV}
\end{equation}
where
\begin{align}
    \begin{split}
        Q_{VO} &= \text{LayerNorm}(X)\\
        K^i_{OV} &= (U\Sigma)^T\\
        V^i_{OV} &= V^T
    \end{split}
\end{align}
\\
The left singular vectors, associated with the columns of $U \Sigma \in \mathbb{R}^{d \times d}$, act as a library of ``keys" $K^i_{OV}$ against which the normalized features $X_l \in \mathcal{H}_S$ are compared.
While the corresponding right singular vectors, associated with rows in $V^T \in \mathbb{R}^{d \times d}$, act as the output values $V^i_{OV}$ that define the direction in which to update the information in the residual stream for a given key.
This interpretation is motivated by the results of \citet{beren2022svd}, where it is shown that the right singular vectors $V^T$ of the $W_{VO}$ matrix tend to have interpretable meanings when decoded using $W_E$, with some of the transformer heads consistently representing a single topic in most of their singular vectors.
We would also like to highlight that, similar to the $W_{QK}$ matrix, the $W_{OV}$ matrix has at least $d-k$ singular values equal to zero.
This means that multiple queries $Q_{VO}$ will map to the zero vector and thus won't update the information in the residual stream, allowing the model to skip the update process if necessary.

\textbf{Note on bias terms:} If the value projection in \autoref{eq:qkv_matrices} contains a bias term $\beta_v \in \mathbb{R}^{1 \times k}$, and the output projection in \autoref{eq:multihead_original} contains a bias term $\beta_o \in \mathbb{R}^{1 \times d}$.
The layer update in \autoref{eq:layer_update} can be re-written as follows:
\begin{equation}
    \label{eq:layer_update_with_bias}
    X_{l+1} = X_l + \beta_o + \sum_{i}^h A^i X^i_{value} + \beta_v {W^i_O}^T
\end{equation}
\\
Here, the term $\beta_v {W^i_O}^T  \in \mathbb{R}^{1 \times d}$ is a bias on the update direction of head $i$, while $\beta_o \in \mathbb{R}^{1 \times d}$ acts as a bias on the entire layer's update.

\subsection{The Feed Forward Module}\label{sec:ff_module}

We use the same interpretation for the feed-forward module as \citet{geva2022transformer, geva2021transformer}.
In it, the feed-forward module behaves similarly to the $W_{VO}$ matrix in the sense that it also acts as a key-value store \citep{geva2021transformer} that proposes directional updates for features in the residual stream \citep{geva2022transformer}.
Similar to the previous section, we will begin by considering the update step after the feed-forward layer:
\begin{equation}
        X_{l+1} = X_l + \text{FeedForward}(\text{LayerNorm(X)})
\end{equation}
where
\begin{equation}
    \text{FeedForward}(X) = f(XW_1)W_2 + \beta_{f}
\end{equation}
\\
\citet{geva2021transformer} note that, for a hidden dimension $d_{hidden}$, the feed-forward module weights $W_1 \in \mathbb{R}^{d \times d_{hidden}}$ and $W_2 \in \mathbb{R}^{d_{hidden} \times d}$ act as key and value matrices, plus a bias term $\beta_f \in \mathbb{R}^{1 \times d}$.
They propose an alternative formulation of the feed-forward layer:
\begin{equation}
    \label{eq:feed_forward_sum}
    \text{FeedForward}(X) = \sum_i^{d_{hidden}} f(X k_i) \cdot v_i + \beta_f = \sum_i^{d_{hidden}} m_i \cdot v_i + \beta_f
\end{equation}
where
\begin{align}
\begin{split}
    k_i &= W_{in}^T[:, i] \quad \in \mathbb{R}^{d \times 1}\\
    v_i &= W_{out}[i, :]  \quad \in \mathbb{R}^{1 \times d}
\end{split}
\end{align}
Such that $W_1$ acts as a storage matrix for keys $k_i$, $W_2$ acts as a storage matrix for values $v_i$, and the activation function $f$ assigns a weight $m_i$ to each value $v_i$ depending on the input $X$.
In \citep{geva2021transformer}, the top-n examples in the training dataset that resulted in the highest $m_i$ coefficients showed interpretable patterns, such that each key $k_i$ in a 16-layer transformer model trained on WikiText-103 \citep{merity2016pointer} was associated with either a syntactical or semantical pattern by human experts.
For the values $v_i$, their role is better understood in terms of the impact that they have on the residual stream (referred to as sub-updates):
\begin{equation}
    \label{eq:layer_update_feed_forward}
    X_{l+1} = X_l + \beta_f + \sum_i^{d_{hidden}} m_i \cdot v_i
\end{equation}
It can be seen that each value $v_i$ modifies the residual stream independently, implying that these share the same space as $W_E$.
Indeed, experiments from \citet{geva2021transformer, geva2022transformer} have shown that many values $v_i$ are semantically meaningful and can be intervened for applications like zero-shot toxic language suppression.

To conclude this subsection, we highlight that the attention and feed-forward modules behave very similarly (see \autoref{eq:layer_update_with_bias} and \autoref{eq:layer_update_feed_forward}), as both calculate relevance scores and aggregate sub-updates for the residual stream.
However, the way the scores and updates are calculated is very different.
The attention module relies primarily on dynamic context for its scores and values, while the feed-forward module relies on static representations.
\subsection{The Word Embedding Matrix \texorpdfstring{$W_E$}{we} and Output Probabilities}

Once all the attention and feed-forward updates have been applied, the output probabilities of the network can be obtained as follows \cite{xiong2020layer}:
\begin{equation}
    \label{eq:out_probs}
    p(Y) = \text{softmax}\big( \text{LayerNorm}(X_L) W_E^T \big)
\end{equation}
\\
In the case where layer normalization has no trainable parameters, \autoref{eq:out_probs} can be interpreted as measuring the similarity between the final layer representation $X_L$ when projected to $\mathcal{H}_S$, and each of the embedding vectors in $W_E$.
Given that all vectors in the projection have the same norm $\sqrt{d}$, the only relevant factor in deciding the output probability distribution $p(y^t) \in [0, 1]^{|V|}$, at a given timestep $t$, is the location of its corresponding vector $x^t_l$ within $\mathcal{H}_S$.
This behavior is very similar to that described by the von Mises-Fisher distribution \citep{fisher1953dispersion}, as both represent distributions parameterized by a reference vector within a hyper-sphere.
Nonetheless, in the case of transformers, the support of the distribution is defined over a discrete set of vectors in $\mathbb{R}^d$, instead of $\mathcal{H}_S$ as a whole.

In the case the layer normalization includes scaling and bias parameters $\gamma$ and $\beta$, the output probabilities are calculated as follows:
\begin{equation}
    \label{eq:out_probs_with_bias}
    p(Y) = \text{softmax} \big( \hat{X_L} \Gamma W_E^T + \beta W_E^T \big)
\end{equation}
where $\hat{X_L}$ is the projection of $X_L$ to $\mathcal{H}_S$ and $\Gamma$ is a diagonal matrix such that $\Gamma_{ii} = \gamma_i$.
The effect of $\Gamma$ on the representation is that of transforming $\mathcal{H}_S$ into an ellipsoid (see Top Right section of \autoref{fig:w_qk}) while $\beta W_E^T$ acts as a bias that assigns higher probability to certain tokens independent of the input.

In both cases (with and without bias and scale parameters), this perspective aligns with that of iterative refinement \citep{jastrzkebski2017residual} discussed in \citet{nostalgebraist2020interpreting, elhage2021mathematical, geva2022transformer, belrose2023eliciting}, given that intermediate representations $X_l$ can always be converted into output probabilities using \autoref{eq:out_probs}.

To conclude this section, we would like to highlight that, by considering the role of layer normalization and how it constrains the representation space, we can get a geometric intuition behind iterative refinement.
We provide a visual interpretation of this concept in \autoref{fig:traveling}.

%% file: 04exps.tex
\section{Experiments}

This section presents our experimental results.
All experiments use pre-trained weights from the 124M parameter version of GPT-2 \citep{radford2019language, karpathy2023nanoGPT} unless stated otherwise.
Code to replicate all experiments is available at: \url{https://github.com/santiag0m/traveling-words}.

\subsection{Impact of Layer Normalization on the Word Embeddings}

To measure the impact of layer normalization on the position of the embedding vectors in $W_E$, we calculated both the $\ell _2$ and cosine distances between the layer-normalized weights and the following settings:
\begin{itemize}
    \item Original: The original word embeddings without any modification
    \item Centered: Original + centering around the mean $\mathrm{E}[w_e]$
    \item Scaled: Original divided by the average vector norm $\mathrm{E}[||w_e||_2]$ and multiplied by $\sqrt{d}$
    \item Centered + Scaled: Original + centering + scaling
\end{itemize}

\begin{table}
    \centering
    \caption{Distance between the normalized embeddings $\text{LayerNorm}(W_E)$ and different transformations of the embedding matrix $W_E$.}
    \vspace{0.3cm}
    \label{tab:e_norm}
    \resizebox{\linewidth}{!}{
        \begin{tabular}{l c c}
            \hline
            \textbf{Setting}                & \textbf{Mean $\ell _2$ Distance} & \textbf{Mean Cosine Distance} \\
            \midrule
            Original                        & 23.747 (0.432)                   & \num{<0.001} (\num{<0.001})   \\
            Centered                        & 24.872 (0.432)                   & 0.150 (0.035)                 \\
            Scaled by $\sqrt{d}$            & 2.413  (1.862)                   & \num{<0.001} (\num{<0.001})   \\
            Centered + Scaled by $\sqrt{d}$ & 14.591  (1.469)                  & 0.150 (0.035)                 \\
            \hline
        \end{tabular}
    }
\end{table}
The results in \autoref{tab:e_norm} show that the mean cosine distance between the original word embeddings and the embeddings after normalization is close to zero, meaning that projection onto $\mathcal{H}_S$ does not modify the orientation of the embedding vectors.
The results also confirm this when centering is applied, as the cosine distance increases significantly when the original vectors are displaced from the origin towards the mean.
On the other hand, it can be seen that the $\ell _2$ distance is high for all settings except for when scaling is applied without centering.
Given an average norm of $\mathrm{E}[w_e]=3.959$ and for $\sqrt{d}=27.713$ we can conclude that the original word embeddings lie between the origin and $\mathcal{H}_S$ rather than on its surface, with different embeddings having different norms.

Variance in the norm of embedding vectors is likely to be a result of the use of the word embedding matrix as a classification layer later in the network (see \autoref{eq:out_probs_with_bias}).
To verify whether this is the case, we select the top and bottom 5 embedding vectors based on the three following criteria:
\begin{itemize}
    \item Norm: The norm of the original embedding vector in $W_E$
    \item Scaled Norm: The norm of the embedding vector when scaled by the Layer Norm parameter $\Gamma$
    \item Norm + Bias: The norm of the original embedding vector plus the bias scores obtained from $\beta W_E^T$
    \item Scaled Norm + Bias: The sum between the Scaled Norm and the bias scores.
\end{itemize}
\begin{table}
    \centering
    \caption{
        Top 5 and Bottom 5 tokens from the word embedding matrix.
        Tokens were sorted according to the relevance of their corresponding embedding vectors under different measurement settings.
    }
    \vspace{0.3cm}
    \label{tab:token_scores}
    \resizebox{\linewidth}{!}{
        \begin{tabular}{c c c c c}
            \hline
            \textbf{Position} & \textbf{Norm}                        & \textbf{Scaled Norm}                                   & \textbf{Norm + Bias} & \textbf{Scaled Norm + Bias} \\
            \hline
            Top 1             & SPONSORED                            & \textbackslash xa9\textbackslash xb6\textbackslash xe6 & ,                    & the                         \\
            Top 2             & \textbackslash x96\textbackslash x9a & tremend                                                & the                  & ,                           \\
            Top 3             & soDeliveryDate                       & \textbackslash x96\textbackslash x9a                   & .                    & and                         \\
            Top 4             & enegger                              & senal                                                  & and                  & a                           \\
            Top 5             & Reviewer                             & millenn                                                & -                    & in                          \\
            \midrule
            Bottom 5          & for                                  & -                                                      & \textbackslash xc0   & \textbackslash x07          \\
            Bottom 4          & an                                   & (                                                      & \textbackslash x07   & \textbackslash x0f          \\
            Bottom 3          & on                                   & ``\textbackslash n"                                    & \textbackslash x10   & oreAndOnline                \\
            Bottom 2          & in                                   & ,                                                      & \textbackslash x11   & \textbackslash x06          \\
            Bottom 1          & at                                   & .                                                      & \textbackslash xfe   & \textbackslash xc1          \\
            \hline
        \end{tabular}
    }
\end{table}
The sorted tokens in \autoref{tab:token_scores} show that considering only the norm of the embeddings is not enough, as tokens that are not commonly used (like `SPONSORED' and `soDeliveryDate') have the highest norms, while common words like `for', `an', `on' and `in' have the lowest norm.
After considering the scaling parameter $\Gamma$, we observe that punctuation signs like the newline character or the comma `,' have the lowest norm, and that there is no clear pattern on the top tokens.
After considering bias, we see that the distribution of top tokens clearly shifts, with punctuation symbols and common words at the top and uncommon bytes at the bottom.
Finally, note that when both scale and bias are considered, the top tokens are consistent with some of the most common words in the English language: `the', `and', `a' and `in' with the only exception being the comma character, which is also very common in natural language, while the bottom tokens are related to uncommon bytes and an anomalous token.

\subsection{Probing Attention Heads with Normalized Representations of Common Nouns}

Next, we use the interpretation from \autoref{sec:w_qk_matrix} and \ref{sec:w_vo_matrix} to probe the attention heads at layers 0, 5 and 11 of the GPT-2 model using as inputs the 100 most common nouns taken from the Corpus of Contemporary American English (COCA) \citep{davies2010corpus}.
First, we transform the embedding matrix $W_E$ according to the normalization parameters specific to each layer (see \autoref{fig:w_qk}) and then multiply the normalized embeddings $W_E^{norm}$ by either $W_{QK}$ or $W_{VO}$.

Then, we perform decoding steps specific to each matrix after multiplication:
\begin{itemize}
    \item For $W_{QK}$, we retrieve the top-k closest embedding vectors from $W_E^{norm}$ based on dot product similarity.
    \item For $W_{VO}$, we add the head-specific and layer-specific output biases (see \autoref{eq:layer_update_with_bias}) to obtain the ``update vectors".
          These update vectors are then added to the original embeddings from $W_E$ and transformed according to the normalization parameters from the last layer; then, we retrieve the top-k closest embeddings from the original $W_E$ based on dot product similarity.
\end{itemize}

\subsubsection{Query-Key Transformations}
In \autoref{tab:qk_layer_0}, we present the results for the Query-Key transformations at layer 0 given the query inputs `time', `life' and `world'.
We note that some of the heads preserve the meaning of the query, as is the case for heads 1, 5 and 10, possibly looking for repetition, while others look for keys that precede it.
Such precedence heads might help to disambiguate the meaning of the words, with examples like: `Showtime' vs. `spacetime', `battery life' vs. `wildlife' and `underworld' vs. `Westworld'.
Other heads appear to be looking for contextual associations, as is the case for head 2, which seems to relate `world' with dates and concepts from the First and Second World wars.
When looking at deeper layers (as shown in \autoref{tab:qk_layer_5} \& \ref{tab:qk_layer_11}), we were not able to identify any meaningful patterns on the query transformations, suggesting that these layers might look for more complex patterns.
\begin{table}
    \centering
    \caption{Transformation of Queries Across Transformer Heads at Layer 0}
    \vspace{0.3cm}
    \label{tab:qk_layer_0}
    \resizebox{\linewidth}{!}{
        \begin{tabular}{l|l|l|l}
            \toprule
                          & \multicolumn{3}{c}{\textbf{Query} $\rightarrow$ \textbf{Keys}}                                                     \\
            \cline{2-4}
            \textbf{Head} & time                                                           & life                      & world                 \\
            \midrule
            0             & Level, [?], offenders                                          & battery, Battery, Battery & legraph, Vers, Malf   \\
            1             & time, time, Time                                               & Life, life, life          & World, world, world   \\
            2             & cinematic, Priest, priest                                      & Notre, fetal, abortion    & 1914, Churchill, 1916 \\
            3             & space, lunch, mid                                              & augh, ertain, ough        & under, Nether, Fort   \\
            4             & soft, heavy, tool                                              & Middle, Hans, Middle      & ether, Unt, Know      \\
            5             & time, time, Time                                               & life, Life, Life          & world, World, world   \\
            6             & Rated, chirop, u                                               & Fukushima, chirop, ulic   & ipt, u, Meta          \\
            7             & Show, bed, Movie                                               & pro, wild, Wild           & Disc, West, West      \\
            8             & java, framework, watch                                         & shark, sharks, Wild       & edit, "\$:/, movie    \\
            9             & stones, pal, cards                                             & Trojan, malware, Wi       & Rogers, COUNTY, Rd    \\
            10            & time, time, Time                                               & life, life, Life          & world, world, World   \\
            11            & Wine, a, food                                                  & PHI, everal, Span         & agus, true, `,'       \\
            \bottomrule
        \end{tabular}
    }
\end{table}

\subsubsection{Key-Value Transformations}
In \autoref{tab:kv_layer_0}, we present the results for the Key-Value transformations for the same three inputs.
For most heads at layer 0, the meaning of the input key is kept as is. However, when the sum of all the heads is considered, we see a slight shift in the meaning of the words.

For heads at layer 5 (shown in \autoref{tab:kv_layer_5}), we see that although most of the heads preserve the meaning of the input keys `life' and `world' (and around half of the heads for the input `time'), the sum of all heads does change the word meaning dramatically, and without a clear output pattern.
As our experiment is limited to testing a single input key at a time, it might be possible that updates in this layer rely more heavily on the composition between multiple keys, which we did not capture.

Finally, in the last layer (\autoref{tab:kv_layer_11}), we see that most individual heads map to seemingly arbitrary values, with only a few preserving the meaning of the input key.
However, when the sum of the heads is considered, the layer preserves the meaning of the input keys.
To test the hypothesis that meaning-preserving heads dominated the layer update, we measured the norm of the output values for each head (before adding the layer-specific bias $\beta_o$).
We found that, in most cases, these heads do not have higher norms.
Instead, heads promoting common tokens like `the', `,' and `and' had the highest norms.
These results suggest that contrary to our hypothesis, the heads at the last layer work together to preserve the meaning of the input keys and mitigate the network's bias towards common tokens.

\begin{table}
    \centering
    \caption{Transformation of Keys Across Transformer Heads at Layer 0}
    \vspace{0.3cm}
    \label{tab:kv_layer_0}
    \resizebox{\linewidth}{!}{
        \begin{tabular}{l|l|l|l}
            \toprule
                          & \multicolumn{3}{c}{\textbf{Key} $\rightarrow$ \textbf{Values}}                                                          \\
            \cline{2-4}
            \textbf{Head} & time                                                           & life                        & world                    \\
            \midrule
            0             & time, Time, time                                               & life, choice, senal         & world, World, worlds     \\
            1             & time, TIME, time                                               & life, lihood, life          & world, Goes, ship        \\
            2             & time, [?], Minutes                                             & life, Life, life            & world, world, World      \\
            3             & time, Time, theless                                            & life, Life, life            & world, World, worlds     \\
            4             & time, time, Time                                               & life, Life, Life            & world, World, world      \\
            5             & time, Time, Time                                               & life, Life, Life            & world, World, worlds     \\
            6             & time, time, Time                                               & life, life, Life            & world, world, Feather    \\
            7             & time, eless, times                                             & life, Experience, Life      & world, World, Abyss      \\
            8             & time, iversary, melodies                                       & life, challeng, conservancy & world, worlds, droid     \\
            9             & time, time, recall                                             & [?], local, Main            & [?], world, local        \\
            10            & equivalents, igation, planes                                   & life, ento, planner         & world, ento, Tanzania    \\
            11            & time, Time, Time                                               & life, Life, +++             & world, World, Trials     \\
            \midrule
            Sum           & time, etime, watch                                             & Indigo, life, crew          & world,  Unleashed, World \\
            \bottomrule
        \end{tabular}
    }
\end{table}

\subsection{Singular Value Decomposition of the \texorpdfstring{$W_{VO}$}{WVO} matrix}

To verify whether the key-value interpretation of $W_{VO}$ matrix proposed in \autoref{sec:w_vo_matrix} is correct, we probe each of its singular vectors (as proposed in \cite{beren2022svd}).
For the left singular vectors $U$ (scaled by $\Sigma$), we use the normalized embeddings $W_E^{norm}$ as a probe, while for the right singular vectors $V^T$, we use the original embeddings $W_E$.
Given that all singular values are constrained to be positive, we get two possible singular vector pairs corresponding to each singular value: $(u, v)$ and $(-u, -v)$.
For ease of analysis, we choose the signed pair with its $v$ component closest to any of the embeddings $w_e \in W_E$, using the dot product similarity.

We did not observe any interpretable pattern for the attention heads at layer 0 and found only one interpretable head at layer 5 (head 10), which referred to terms in politics and chemistry.
However, we found that most heads in layer 11 were interpretable (except for heads 5, 7 and 9) and present the results for all heads in \autoref{sec:svd_heads}.
An illustrative case of these patterns is head 3, where most of its singular vector mappings are related to jobs or industries.
For example, `Dairy' maps to `USDA' (the United States Department of Agriculture), `engine' to `drivers', `trading' to `Sales' and so on.
Similar patterns were present in other heads, listed as follows:
\begin{itemize}
    \item Head 0: Formatting and punctuation symbols (end of text, new line, brackets and parenthesis)
    \item Head 1: Gender words
    \item Head 2: Proper Nouns (Places)
    \item Head 3: Jobs / Industries
    \item Head 4: Letters and Numbers
    \item Head 6: Suffixes and Prefixes related to the ending and beginning of words
    \item Head 8: Punctuation symbols
    \item Head 10: Proper Nouns (First and Last names)
    \item Head 11: The identity function (input similar to the output)
\end{itemize}

We found that these patterns were consistent with those obtained in the ``Key $\rightarrow$ Value" results from \autoref{tab:kv_layer_11}, implying that the subject-specific behavior of the singular vectors is reflected in the input-output transformations of the attention heads.
These results complement previous work from \citet{beren2022svd}, in which only the right singular vectors $V^T$ were considered.

\subsubsection*{SVD of the \texorpdfstring{$W_{QK}$}{WQK} matrix}

In additional experiments on the SVD of the $W_{QK}$ matrix, we found that some singular vector pairs had clear associations.
For example, in head 0 of layer 0, we found some associations related to programming languages (`self, class, =, import' $\rightarrow$ `Python') and digital cameras (`Video, 264, minutes' $\rightarrow$ `Nikon, lineup, shot, camera') but we could not identify any specialization for the heads.
Surprisingly, we did find that heads at the last layer had identifiable patterns on their left singular vectors (associated with the queries) consistent with those listed for the $W_{VO}$ matrix (punctuation for head 0, gender for head 1, and so on), but no clear patterns were identified for the right singular vectors.

\subsection{Visualizing Iterative Refinement}

\begin{figure}[ht]
    \centering
    \includegraphics[width=0.8\linewidth]{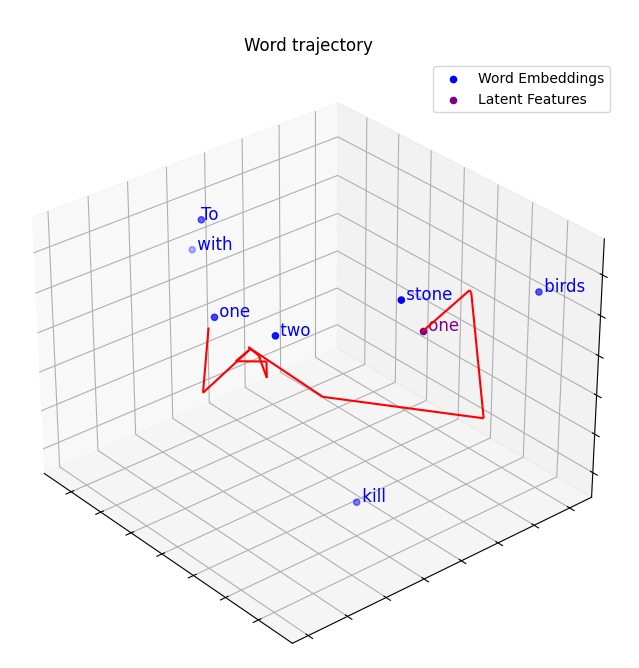}
    \caption{
        UMAP 3D projection of the phrase `To kill two birds with one stone'.
        The original word embeddings are in blue, the final latent representation for the second-to-last token (`one') in purple, and its trajectory in red, with each trajectory segment representing an update between transformer blocks.
        Note that the latent representation starts close to its corresponding embedding, `one', and gets closer to that of the next token, `stone', with each update.
    }
    \label{fig:trajectory}
\end{figure}

Finally, we visualize how the information in the residual stream is updated (i.e. the iterative refinement process) leveraging dimensionality reduction techniques, as shown in \autoref{fig:trajectory}.
For this, we chose the test sentence `To kill two birds with one stone', as the predictability of its last token, `stone', given the previous context was high (correctly predicted by the model) and none of the words in the sentence repeated.
To project the high dimensional embeddings into 3D space, we used UMAP \citep{mcinnes2018umap}, with Laplacian Eigenmap initialization \citep{belkin2001laplacian, kobak2021initialization}, and we fit the transform using the first 10,000 embedding vectors from $W_E$ to accurately reflect proximity in the original embedding space.
We show the original embedding tokens as reference (in blue) and plot the trajectory of the second-to-last token, `one', as we process the entire sequence (with added positional embeddings) throughout the network.
For each layer, we transform the latent representations in the residual stream using the normalization parameters from the final output layer before projecting with UMAP.
It can be seen that the representation of the second-to-last token shifts from its original meaning (`one') towards the meaning of the next token (`stone').
Although the figure also shows the magnitude and direction of each update in the trajectory, it is important to mention that these quantities might have been modified due to the dimensionality reduction process.

%% file: 05conclusion.tex
\section{Conclusion}

We have presented a new interpretation of transformer models based on the geometric intuition behind each component and how all these components come together as the transformation of the meaning of one input token to the next.

First, we showed how layer normalization can be better understood as a projection of latent features in $\mathbb{R}^d$ to a $(d-1)$-dimensional hyper-sphere and provide experimental evidence that the word embeddings learned by GPT-2 are distributed toward different directions of the hyper-sphere.
We also showed that the parameters of the final normalization layer are crucial to obtain high-scoring tokens consistent with high-frequency tokens in the English language.

Next, we discussed the role of the $W_{QK}$ and $W_{VO}$ matrices as transformations related to the hyper-sphere, with $W_{QK}$ as an affine transformation that overlaps queries and keys, and $W_{VO}$ as a key-value map between the hyper-sphere and the original embedding space.
These intuitions were tested with probing experiments, showing promising results in understanding the role of query-key attention in earlier layers and extending the results from \citet{beren2022svd} on the subject-specific nature of the $W_{VO}$ matrix in attention heads at deeper layers.

Finally, we integrated these ideas and the impact of each component on the residual stream to provide visual evidence on how the iterative refinement process works within transformers.

%% file: 06appendix.tex
\appendix
\renewcommand{\thesection}{\Alph{section}}  %
\renewcommand{\thefigure}{\Alph{section}.\arabic{figure}}  %
\setcounter{figure}{0}  %

\setcounter{table}{0}
\renewcommand{\thetable}{\Alph{section}.\arabic{table}}

\section{Attention Head Transformations for Layers 5 and 11}

\subsection{Query-Key Transformations}

\begin{table}[h]
\centering
\caption{Transformation of Queries Across Transformer Heads at Layer 5}
\vspace{0.3cm}
\label{tab:qk_layer_5}
\begin{adjustbox}{width=1.2\linewidth,center}
\begin{tabular}{l|l|l|l}
\toprule
& \multicolumn{3}{c}{\textbf{Query} $\rightarrow$ \textbf{Keys}} \\
\cline{2-4}
\textbf{Head} & time & life & world \\
\midrule
0 & depend, annot, reason & so, inf, char & Lab, dev, Dev \\
1 & they, themselves, Vers & they, Im, depend & come, once, haven \\
2 & Nepal, ":[", —" & `….', `…', Home & posted, Logged, ideologically \\
3 & appeared, actually, had & posted, axle, .avascript & aryl, Ala, GA \\
4 & attract, CP, contained & misconception, (?, trophy & separatists, activists, extremists \\
5 & Plum, rice, Vers & Sniper, too, hides & Prim, Bright, am \\
6 & en, annually, – & following, Generator, Library & §§, tournaments, StarCraft \\
7 & Wis, def, individual & y, ier, od & Af, Gh, agle \\
8 & condition, intensive, inf & prol, operation, splend & Ard, marketplace, dev \\
9 & post, market, destinations & She, steal, etc & strategy, pd, budget \\
10 & jugg, continuously, Center & essim, enter, tast & exploration, jugg, PLAY \\
11 & straight, interview, fucking & --, Eva, related & spotlight, television, TV \\
\bottomrule
\end{tabular}
\end{adjustbox}
\end{table}

\begin{table}[h]
\centering
\caption{Transformation of Queries Across Transformer Heads at Layer 11}
\vspace{0.3cm}
\label{tab:qk_layer_11}
\begin{adjustbox}{width=1.2\linewidth,center}
\begin{tabular}{l|l|l|l}
\toprule
 & \multicolumn{3}{c}{\textbf{Query} $\rightarrow$ \textbf{Keys}} \\
\cline{2-4}
\textbf{Head} & time & life & world \\
\midrule
0 & UNCLASSIFIED, opausal, ster & opausal, backstage, piece & routine, cat, ocular \\
1 & assion, upp, pir & pir, Virgin, appa & Frontier, theater, onies \\
2 & heid, GI, rict & heid, apy, brance & region, urgy, encyclopedia \\
3 & opic, href, Hitchcock & susceptibility, space, opic & league, space, opic \\
4 & gy, lots, whatever & his, whichever, gy & whichever, whatever, underworld \\
5 & olesterol, tx, erc & olesterol, avy, iana & wealth, Digest, Market \\
6 & ones, volatile, RIS & volatile, olesterol, idency & wealth, useum, theatre \\
7 & whichever, ivalent, lower & mortal, whichever, living & -\$, complex, world \\
8 & ove, HTTP, metaphysical & spiritual, metaphysical, bio & Endless, metaphysical, Marvel \\
9 & Productions, actic, fare & stuff, ience, Productions & entertainment, stuff, World \\
10 & -, code, ing & -, core, ola & core, Labs, ourse \\
11 & emb, ivan, Union & Tour, etc, iona & pires, si, Tour \\
\bottomrule
\end{tabular}
\end{adjustbox}
\end{table}

\newpage

\subsection{Key-Value Transformations}

\begin{table}[h]
\centering
\caption{Transformation of Keys Across Transformer Heads at Layer 5}
\vspace{0.3cm}
\label{tab:kv_layer_5}
\begin{adjustbox}{width=1.2\linewidth,center}
\begin{tabular}{l|l|l|l}
\toprule
 & \multicolumn{3}{c}{\textbf{Key} $\rightarrow$ \textbf{Values}} \\
\cline{2-4}
\textbf{Head} & time & life & world \\
\midrule
0 & BuyableInstoreAndOnline, [?], time & life, advertising, Life & world, opathy, qus \\
1 & MON, Sophia, time & mallow, cause, unn & world, Cav, fect \\
2 & time, qualified, understatement & life, life, Life & world, World, auri \\
3 & )?, >), ?' & \textbackslash] $=>$, life, \textbackslash \textbackslash" $>$ & world, \textbackslash] $=>$, \%" \\
4 & time, TIME, Sabha & Izan, eworld, ieu & world, izons, orld \\
5 & destro, time, rall & life, Life, agre & world, toget, enthusi \\
6 & time, time, TIME & life, Life, life & world, World, WORLD \\
7 & time, corrid, patch & life, Life, life & world, mathemat, redes \\
8 & NetMessage, [?], ibu & venge, idth, aten & ULTS, Magikarp, [?] \\
9 & [?], [?], amina & raviolet, los, SPONSORED & Kraft, quickShipAvailable, Berks \\
10 & time, contrace, Symphony & life, Life, life & world, World, worlds \\
11 & otle, ide, Ide & framing, plot, plots & ittee, rf, pawn \\
\midrule
Sum & externalActionCode, ]), issance & ahon,  awa, ]" & Magikarp, Hig, ETHOD\\
\bottomrule
\end{tabular}
\end{adjustbox}
\end{table}

\begin{table}[h]
\centering
\caption{Transformation of Keys Across Transformer Heads at Block 11}
\vspace{0.3cm}
\label{tab:kv_layer_11}
\begin{adjustbox}{width=1.2\linewidth,center}
\begin{tabular}{l|l|l|l}
\toprule
 & \multicolumn{3}{c}{\textbf{Key} $\rightarrow$ \textbf{Values}} \\
\cline{2-4}
\textbf{Head} & time & life & world \\
\midrule
0 & \textbackslash n,  ", `` &  \{, >, "\# &  [, [* ,[ \\
1 & player, party, Party & youth, House, Youth & party, Trump, party \\
2 & Lisp, Ö, ¨ & [?], Quincy, Yemen & Scotland, Osborne, Scotland \\
3 & Weather, cinem, weather & life, euth, Life & world, Worlds, geop \\
4 & b, k, 2 & inav, d, 4 & i, V, Rivals \\
5 & Part, Show, part & Well, Well, saw & sees, works, View \\
6 & Sub, AM, BR & West, West, East & Sub, Under, ob \\
7 & Journal, Air, Online & home, Home, house & home, Home, internet \\
8 & `,', the, and & `,', the, and & the, `,', and \\
9 & interaction, impression, experience & encounter, belief, encounters & reservations, Illusion, illusions \\
10 & time, TIME, Time & life, life, LIFE & world, world, worlds \\
11 & time, time, Time & life, LIFE, life & world, oy, door \\
\midrule
Sum & time,  Time,  time & life,  Life, Life & world,  Worlds,  worlds\\
\bottomrule
\end{tabular}
\end{adjustbox}
\end{table}

\newpage

\section{\texorpdfstring{$W_{VO}$}{WVO} SVD per Head for Layer 11}
\label{sec:svd_heads}

\renewcommand{\arraystretch}{1.3}

\begin{table}[h]
\centering
\caption{Left and Right Singular Vectors at Layer 11 - Head 0}
\label{tab:svd_head_0}
\resizebox{0.94\linewidth}{!}{
\begin{tabular}{c|p{0.4\linewidth}|p{0.4\linewidth}}
\toprule
\textbf{Rank} & \textbf{Top-3 Left Words} & \textbf{Top-3 Right Words} \\
\midrule
0 &  shenan,  cryst,  encount & DragonMagazine, ertodd, soDeliveryDate \\
1 &  another, Iv, sil &  trave, BuyableInstoreAndOnline,  convol \\
2 &  Sebastian,  Luke,  humankind & quickShipAvailable, EStream, MpServer \\
3 & rans,  thereby,  hem & BuyableInstoreAndOnline,  acknow, Buyable \\
4 & sectional,  [+],  Winged & ThumbnailImage, \textbackslash ufffd\textbackslash ufffd\textbackslash u58eb, Orderable \\
5 & abl, isc, Ah & etheless, olson, llah \\
6 & \textless\textbar endoftext\textbar\textgreater, Advertisements, cest & \textless\textbar endoftext\textbar\textgreater, Advertisements,  kindred \\
7 & ococ, ilan,  guest &  pard,  MBA, uid \\
8 & ]., ],, ]; &  [,  [*, [ \\
9 & \textbackslash n\textbackslash n, ),, cakes & \textbackslash n\textbackslash n, Quote,  Quote \\
10 &  snaps,  Bills,  Texans &  lineback,  Chargers,  Packers \\
11 & )..., ...), )." &  (\textbackslash u00a3,  (,  (?, \\
12 & pen, cle,  Orioles & ."",  [, .") \\
13 & ](, drm, Updated & \textbackslash n\textbackslash n, [/,  [/ \\
14 &  RBI,  Field,  Triple &  RHP,  RBI,  Negro \\
15 & pod, illus,  Maple & ipeg,  aboriginal,  "\textbackslash u2026 \\
16 &  am, 'm, hearted & SPONSORED, Newsletter,  .... \\
17 & Document,  whit, Scott & SPONSORED, tsky,  Ras \\
18 & gen, idd, anned &  Ukrain,  prin,  rul \\
19 &  Ryder, icz, abet & istries, plet,  Gad \\
\hline
\end{tabular}
}
\end{table}

\begin{table}[h]
\centering
\caption{Left and Right Singular Vectors at Layer 11 - Head 1}
\label{tab:svd_head_1}
\resizebox{\linewidth}{!}{
\begin{tabular}{c|p{0.4\linewidth}|p{0.4\linewidth}}
\toprule
\textbf{Rank} & \textbf{Top-3 Left Words} & \textbf{Top-3 Right Words} \\
\midrule
0 &  Customers,  However,  Customer & \textbackslash u899a\textbackslash u9192, natureconservancy, racuse \\
1 &  mint,  Anne,  Marie &  hers,  actress,  Denise \\
2 & ook,  Child, ooks & parents, Parents, Children \\
3 & gow, abad,  BEL &  boy, student,  Guy \\
4 & eries,  girl,  girls &  Girl,  girl,  Queen \\
5 &  Marie,  Sue,  Patricia &  Woman, woman,  woman \\
6 &  Him, les, LCS &  Person,  Persons, Person \\
7 & ndra,  Joint, rity & Her,  Her,  femin \\
8 &  Coach,  recapt,  Players &  Players,  Coach,  coaches \\
9 & istries, WAYS, INAL & god,  Allaah,  God \\
10 &  Ens,  offspring,  statute & male,  males, Woman \\
11 &  Junction, hole,  Abdullah &  girl,  daddy,  Neighbor \\
12 & HR,  ig, akings &  Major, Major,  minors \\
13 &  reunion,  Madison, mes &  boys,  males, Girls \\
14 & asting, uba, ynt &  mom,  moms,  Jim \\
15 & ately, ynam, OUS &  doctoral,  apprentice, Child \\
16 & ifier, Come,  Weekly &  class,  owners, Class \\
17 &  Confederation, ATE,  ingredient & Students,  Students,  Ms \\
18 & athon, jen,  candidates &  Candidate,  candidate,  traveler \\
19 &  Pres, ently, Secure &  character, Characters, Character \\
\hline
\end{tabular}
}
\end{table}

\begin{table}[h]
\centering
\caption{Left and Right Singular Vectors at Layer 11 - Head 2}
\label{tab:svd_head_2}
\resizebox{\linewidth}{!}{
\begin{tabular}{c|p{0.4\linewidth}|p{0.4\linewidth}}
\toprule
\textbf{Rank} & \textbf{Top-3 Left Words} & \textbf{Top-3 Right Words} \\
\midrule
0 & orpor,  rul,  Bolivia &  Adelaide,  Edmonton,  Calgary \\
1 & ball, ERY, hem &  Filipino,  Ultron,  ANC \\
2 & \textbackslash u30f3\textbackslash u30b8,  else, Lib & Ruby,  Scarborough,  Erit \\
3 & verb,  Lamar,  Ankara & Detroit,  Detroit,  Wenger \\
4 & iana, amacare, edia &  Zoro,  Shelby,  Tehran \\
5 &  Gw, otle,  Rangers & \textbackslash u00ed,  Jinn,  Texans \\
6 & ration,  Rim, ially &  Yang,  McCain, Yang \\
7 &  detector, OTOS,  Petersen &  Chilean,  Pharaoh, ffen \\
8 & ald, benefit, ahon &  Petersburg,  Henderson,  Kessler \\
9 & scope, whe, verse & acio,  Mits,  Jacobs \\
10 &  Gators,  Laden,  SEAL &  Malfoy,  Swanson, Romney \\
11 &  Lilly, \textbackslash u00e9t, lla &  Greenwood, Collins,  Byrne \\
12 & ister, ority, isters &  Niagara,  Maharashtra, soDeliveryDate \\
13 &  Paulo, nesota,  Clayton &  Loki, \textbackslash u011f,  Finnish \\
14 & creen,  Cron, Base &  Pike,  Krishna,  Satoshi \\
15 & lake, SP, seeing &  Alberta,  Arlington,  McKin \\
16 &  Bowie, ystem, rey &  Bowie, Murray, Utah \\
17 & head, ding, ressed &  Bulgar,  Warcraft,  Crimean \\
18 &  Venom, elman, lyn &  SJ, Brit, Gordon \\
19 & wright, ansas, arta &  NXT,  Metroid,  Aether \\
\hline
\end{tabular}
}
\end{table}

\begin{table}[h]
\centering
\caption{Left and Right Singular Vectors at Layer 11 - Head 3}
\label{tab:svd_head_3}
\resizebox{\linewidth}{!}{
\begin{tabular}{c|p{0.4 \linewidth}| p{0.4 \linewidth}}
\toprule
\textbf{Rank} & \textbf{Left Words} & \textbf{Right Words} \\
\midrule
0 &  suburbs,  restaur,  \textbackslash ufffd & DragonMagazine, BuyableInstoreAndOnline, \textbackslash ufffd\textbackslash u9192 \\
1 &  Dairy,  farm,  Veget &  USDA,  Dairy,  cows \\
2 &  engine,  drivers,  Motor &  Drivers,  drivers,  driver \\
3 &  trading,  trade,  shoppers & Sales,  retailers,  shoppers \\
4 &  instrument,  musical,  guitar &  Billboard,  halftime,  Grammy \\
5 &  sail,  boat,  sailing &  sail,  sailing, autical \\
6 &  teachers,  teacher,  school &  teachers, uberty,  curric \\
7 &  baker,  kindergarten,  bakery &  baker,  SERV,  kindergarten \\
8 &  apparel,  prison,  recruiting &  Sail,  Prison,  jail \\
9 &  shelter,  indoors,  shelters &  shelters,  shelter,  Radiant \\
10 &  tribe,  fish, fish &  dred,  whales,  fisheries \\
11 &  workers,  jobs,  job & workers, worker,  subcontract \\
12 &  Derrick,  tribe,  Tribal & Seg, forest,  Derrick \\
13 &  chess,  Chess,  seating &  Chess,  chess,  Sheldon \\
14 &  Soy,  Satellite,  astronauts &  Soy,  Satellite,  transmissions \\
15 &  Anim,  visa, Imm &  exhib, Anim,  Imm \\
16 &  medicine,  diagnose,  doctors & Doctors, hospital, doctor \\
17 &  boxing,  trainer,  spar &  boxing,  spar, UFC \\
18 &  gun,  firearm,  Sheriff &  ITV,  Decoder,  Geral \\
19 &  gambling,  tournaments,  tournament &  gambling, Gaming,  tournaments \\
\hline
\end{tabular}
}
\end{table}

\begin{table}[h]
\centering
\caption{Left and Right Singular Vectors at Layer 11 - Head 4}
\label{tab:svd_head_4}
\resizebox{\linewidth}{!}{
\begin{tabular}{c|p{0.4\linewidth}|p{0.4\linewidth}}
\toprule
\textbf{Rank} & \textbf{Top-3 Left Words} & \textbf{Top-3 Right Words} \\
\midrule
0 &  them,  their,  him & cloneembedreportprint, \textbackslash u899a\textbackslash u9192,  \textbackslash u30b5\textbackslash u30fc\textbackslash u30c6\textbackslash u30a3 \\
1 & iator, ive, ibur & natureconservancy,  Canaver, \textbackslash u25fc \\
2 & if, born,  forces &  the, ., , \\
3 & ually, ,.,  therein & Buyable,  misunder, lehem \\
4 & irk, struct, actly &  1,  2,  9 \\
5 & uku, handle, eenth &  nineteen,  seventeen,  seventy \\
6 & ensional, insk, ploy &  M, M,  m \\
7 &  allowance, \textbackslash u2605,  ther &  ii, Bs,  B \\
8 & ylon, works, plays &  EDITION, o\textbackslash u011f, nt \\
9 & ysc, oreal,  Friend &  B,  K, B \\
10 & redits, rossover, ameron &  F,  K,  k \\
11 &  Tiger, urses, aught &  N,  W,  C \\
12 & aughter, gling, eland &  L,  l, L \\
13 & othe, cano, ensity &  S,  s, S \\
14 & ISTORY,  hum, pots &  H, H,  h \\
15 & gers, iegel, ki &  S,  s, S \\
16 & ya,  seq, est & selves,  T,  i \\
17 & tl, ictionary,  latch &  R, R,  D \\
18 &  Fres, pine, delay &  R,  u, llah \\
19 &  Shades, went, culosis &  G, G,  S \\
\hline
\end{tabular}
}
\end{table}

\begin{table}[h]
\centering
\caption{Left and Right Singular Vectors at Layer 11 - Head 5}
\label{tab:svd_head_5}
\resizebox{\linewidth}{!}{
\begin{tabular}{c|p{0.4\linewidth}|p{0.4\linewidth}}
\toprule
\textbf{Rank} & \textbf{Top-3 Left Words} & \textbf{Top-3 Right Words} \\
\midrule
0 &  assail,  challeng,  achie & ertodd, \textbackslash u25fc, \textbackslash ufffd\textbackslash u9192 \\
1 & WARE,  padding, req & \textbackslash u9f8d\textbackslash u5951\textbackslash u58eb, StreamerBot, soDeliveryDate \\
2 & uing, anche,  Inquis & heit, MpServer,  partName \\
3 & ward, ops, actory &  builds,  projects, Building \\
4 & ary, bell, vis & ouf, unt,  article \\
5 & ments,  Poo, emo &  Will, Will, terday \\
6 &  abdom,  book,  Til &  reads,  read, writing \\
7 &  admission,  Fighters, agy &  model,  Models, ilib \\
8 & line, lines, se &  line,  lines,  Hold \\
9 & iness, less, ood & udic,  ridden, usky \\
10 &  absence, inar,  Miko &  place, Must, must \\
11 & hawk, nect, aff & esson,  sees, scene \\
12 & ie, een, ennett &  Say, ighting, features \\
13 &  Peaks,  construed, anguages &  finding, find,  Find \\
14 & ming, mers, pling & ufact,  Put, say \\
15 &  Authority, urated,  disregard &  record,  records,  Record \\
16 & cript,  Seen, Crash &  Written, course, arium \\
17 & ually,  gladly, ously & tions, show,  find \\
18 & im, ading,  Expand &  image, Image,  Image \\
19 &  NX, W, ees & swer, \textbackslash u30c7\textbackslash u30a3,  report \\
\hline
\end{tabular}
}
\end{table}

\begin{table}[h]
\centering
\caption{Left and Right Singular Vectors at Layer 11 - Head 6}
\label{tab:svd_head_6}
\resizebox{\linewidth}{!}{
\begin{tabular}{c|p{0.4\linewidth}|p{0.4\linewidth}}
\toprule
\textbf{Rank} & \textbf{Top-3 Left Words} & \textbf{Top-3 Right Words} \\
\midrule
0 & issue, txt, Princ & isSpecialOrderable, DragonMagazine, \textbackslash ufffd\textbackslash ufffd \\
1 & mes,  same, resa &  guiActiveUn,  Yanuk,  Beir \\
2 & eatured, avier, AMES & quickShipAvailable, BuyableInstoreAndOnline, RH \\
3 &  Levine, estone,  Bronx & skirts, Els,  Bris \\
4 & lder, xit, Sav &  Sov,  grap,  Al \\
5 & xual, ss,  soup &  Orient, owship,  toile \\
6 & rices, glers, lishing &  Uni,  Tik,  en \\
7 & imation, hammer, nels &  BAD,  Ze,  sa \\
8 &  saturated, lying,  Past &  Ry,  AG,  Val \\
9 & activity, ozy, oko &  Ay,  AW, Ay \\
10 & ows, aghan, ergy &  Gul, cl,  Nex \\
11 & yrs, ish, hood &  Wh,  Har,  Mart \\
12 & omp,  grandmother, MS &  sidx,  Alb,  CTR \\
13 & ses, ski,  doctor &  AD, ython, Ty \\
14 & heed,  Monthly, angan &  OPS, Tur, Tam \\
15 &  Agency, VP, lex &  Red, Grey, Redd \\
16 & FORE, sil, hing & wcsstore, uci, Winged \\
17 & idences, ining, ahl &  Ste,  Pend,  hal \\
18 & iance,  taxpayers, anches &  Fuj,  appl,  Zamb \\
19 & ischer, apo,  hiatus &  Zamb,  Zer,  Nek \\
\hline
\end{tabular}
}
\end{table}

\begin{table}[h]
\centering
\caption{Left and Right Singular Vectors at Layer 11 - Head 7}
\label{tab:svd_head_7}
\resizebox{\linewidth}{!}{
\begin{tabular}{c|p{0.4\linewidth}|p{0.4\linewidth}}
\toprule
\textbf{Rank} & \textbf{Top-3 Left Words} & \textbf{Top-3 Right Words} \\
\midrule
0 &  shortest, ses,  mentally & iHUD, DragonMagazine, Downloadha \\
1 &  our,  ourselves,  we &  ourselves,  ours,  our \\
2 &  himself, lements,  them & \textbackslash u899a\textbackslash u9192, natureconservancy, ertodd \\
3 & etitive, EStream, workshop & FTWARE, SourceFile, \textbackslash ufffd\textbackslash u9192 \\
4 & \textbackslash u0627\textbackslash u0644, holders,  mileage &  your,  Free,  Your \\
5 &  am, 'm,  myself &  my,  myself,  me \\
6 &  themselves, auder, ighthouse & Companies,  theirs,  THEIR \\
7 & stract, hop, \textbackslash u00a2 & soDeliveryDate, Civil,  civilian \\
8 & bage, ros,  hyster & bage, aukee,  Free \\
9 &  shop, acter,  Shop &  Humans, ourning,  electronically \\
10 &  <+,  myself, pse &  my,  myself,  markets \\
11 & Hold, SE, istant & ilage, roups, usra \\
12 & uffs, VG, GG & verty,  Leilan, Soft \\
13 & sters, ual, ted &  machine, machine,  business \\
14 & making, weights, mare &  centrif, istani,  culture \\
15 & uador, oust, ertain &  us,  ours,  our \\
16 & vable, cam, ophy &  system,  System,  systems \\
17 &  exch, velength, un &  Games, abeth, gaming \\
18 &  latex,  Edwards,  Conway & Commercial,  Community, community \\
19 & ificial, rating, nces & ificial,  System,  technology \\
\hline
\end{tabular}
}
\end{table}

\begin{table}[h]
\centering
\caption{Left and Right Singular Vectors at Layer 11 - Head 8}
\label{tab:svd_head_8}
\resizebox{\linewidth}{!}{
\begin{tabular}{c|p{0.4\linewidth}|p{0.4\linewidth}}
\toprule
\textbf{Rank} & \textbf{Top-3 Left Words} & \textbf{Top-3 Right Words} \\
\midrule
0 &  the,  in,  a & \textbackslash ufffd\textbackslash ufffd\textbackslash ufffd,  guiActiveUn, cloneembedreportprint \\
1 & *., .,  determin & ,,  the, - \\
2 &  and, ,,  Un & arnaev, DragonMagazine, BuyableInstoreAndOnline \\
3 & ?", ?), ?), & ?'", TPPStreamerBot, '," \\
4 & ,'",  GIF, ,'' & Orderable, \textbackslash ufffd, \textbackslash ufffd \\
5 & .'", They, .' & .'", '.", )." \\
6 & ,', \textbackslash u2010, ,'" & ,'",  ',, ,' \\
7 & ,', ,'", ',' & ,'", ,', '," \\
8 & .', ,', '. & .', ,',  '. \\
9 &  her,  she, She & she,  hers,  her \\
10 &  \textbackslash ufffd, \textbackslash ufffd,  `` & \textbackslash ufffd,  \textbackslash ufffd,  `` \\
11 &  "..., ",", ", & )",, ),", "), \\
12 & ).", ")., ..." & ).", ."[, "). \\
13 &  us, ),",  our & ),", ).", .") \\
14 &  ));,  );,  ), &  ",,  ',,  )); \\
15 & ]., ];, ], & \};, ];, '; \\
16 &  ...,  ...", ... & ...], :],  ..." \\
17 & ?], !], .] & \textbackslash u2026], !], ?] \\
18 &  ();,  her, He &  hers,  ();, His \\
19 &  \textbackslash u00ad, \textbackslash u300f,  You & \textbackslash u300f, >.,  \textbackslash u00ad \\
\hline
\end{tabular}
}
\end{table}

\begin{table}[h]
\centering
\caption{Left and Right Singular Vectors at Layer 11 - Head 9}
\label{tab:svd_head_9}
\resizebox{\linewidth}{!}{
\begin{tabular}{c|p{0.4\linewidth}|p{0.4\linewidth}}
\toprule
\textbf{Rank} & \textbf{Top-3 Left Words} & \textbf{Top-3 Right Words} \\
\midrule
0 & esthes, Eat,  pts & DragonMagazine,  Canaver, natureconservancy \\
1 & ups,  motors,  hinted &  confir, \textbackslash ufffd,  unlaw \\
2 &  pursue,  pursuit,  Frie & ticket,  Desire, iferation \\
3 & posted, dates,  rece &  achievement,  unlocking,  Hilbert \\
4 &  differential, prise, ushing &  acceptance, handled,  accepting \\
5 &  Hide, etsu, LET &  optimizations,  prioritize,  emphasized \\
6 & ously, uffer, ca & opsis, \textbackslash u30df, stall \\
7 &  ann,  Horn,  Specifications &  restraint, notice,  surprises \\
8 &  supremacy, argon, ifier &  ACTIONS,  Contin, rue \\
9 & ling, ceived,  inf &  errors,  misunderstanding,  accuracy \\
10 & ittal, ampton, feld &  denotes,  denote,  hazard \\
11 & inf, andy, ery &  plagiar,  mentors,  recommending \\
12 &  Soon, \textbackslash ufffd, \textbackslash ufffd &  lax,  Talks,  Fell \\
13 & cia, war,  Fighters &  dissatisf,  consum,  dissatisfaction \\
14 & NAS,  Schwar, Streamer &  delet,  sidx, inem \\
15 &  Glory, uan, ment & Reviewed,  Congratulations,  congratulations \\
16 & frey,  clay, essional &  quirks,  Integration,  distinguishing \\
17 & uck, marked,  Request &  appreciation,  Guidelines,  guidelines \\
18 & prints,  forcefully,  Cel &  conviction,  convictions,  impressions \\
19 & utic, endez, inging &  disag,  bruising,  spo \\
\hline
\end{tabular}
}
\end{table}

\begin{table}[h]
\centering
\caption{Left and Right Singular Vectors at Layer 11 - Head 10}
\label{tab:svd_head_10}
\resizebox{\linewidth}{!}{
\begin{tabular}{c|p{0.4\linewidth}|p{0.4\linewidth}}
\toprule
\textbf{Rank} & \textbf{Top-3 Left Words} & \textbf{Top-3 Right Words} \\
\midrule
0 & above,  former, dm & \textbackslash u25fc, Downloadha,  Canaver \\
1 &  Cohen, oku,  Corporation & be, ache,  the \\
2 &  liar,  Ross,  Irving &  Rossi,  Mind, Zen \\
3 &  Treatment,  MT,  tubing & etts, Taylor, Tan \\
4 &  Torch, dt,  Honour &  Divinity,  marqu, vine \\
5 & ==,  Sinn,  imitation &  Stafford,  Bradford,  Halo \\
6 & asks, fitted,  caution &  BW, BW,  Berger \\
7 & encer,  hero,  success &  Gon, Johnny,  PATH \\
8 &  Chung, anke, IRE &  Chennai,  Carey,  Carmen \\
9 &  Commodore, iom,  attract &  curry,  Cunningham,  clam \\
10 &  earth, CS, oyal & Sov,  Trin, paralle \\
11 & ramid,  el, DIT &  Hilton,  diarr, \textbackslash ufffd\textbackslash u9192 \\
12 & ulla, alde, uality &  McInt, alde,  Idle \\
13 &  cam, write, ports &  Cave,  Chal,  Connie \\
14 & buf, anne,  Emin &  Dwar,  Dwarf,  Das \\
15 & job, play,  job & buquerque,  Liber, reb \\
16 &  ASC, ector, Order &  Sorceress,  Alic,  Astro \\
17 & ting, enced, te & Forest,  Kan,  tree \\
18 & ater,  Turner, UAL & \textbackslash u9f8d\textbackslash ufffd,  Omn,  Gamma \\
19 &  Matrix, RIP, oping & Fed,  STEP, Rand \\
\hline
\end{tabular}
}
\end{table}

\begin{table}[h]
\centering
\caption{Left and Right Singular Vectors at Layer 11 - Head 11}
\label{tab:svd_head_11}
\resizebox{\linewidth}{!}{
\begin{tabular}{c|p{0.4\linewidth}|p{0.4\linewidth}}
\toprule
\textbf{Rank} & \textbf{Top-3 Left Words} & \textbf{Top-3 Right Words} \\
\midrule
0 & 8, 9, 6 & \textbackslash u899a\textbackslash u9192, cloneembedreportprint, StreamerBot \\
1 & ", [?], , & DragonMagazine, cloneembedreportprint, ertodd \\
2 &  puff, rem,  Ey & \textbackslash ufffd\textbackslash ufffd\textbackslash u58eb, catentry,  Flavoring \\
3 &  air,  compressor,  exchange &  air,  blow,  nose \\
4 &  burn,  burning,  burns &  burns,  burning,  burn \\
5 &  smoke,  blowing,  sky &  smoke,  clouds,  airflow \\
6 &  light,  shade,  lighting &  light,  illumination, Light \\
7 &  break,  breaks, Bre & breaks,  breaker,  broken \\
8 &  finger,  air,  registrations & finger,  finger,  Feet \\
9 &  rolls,  roll,  rolled &  Rolls,  rolls,  Ludwig \\
10 &  opening,  opened,  closing &  opened,  closes,  opening \\
11 & ause,  blank,  generating & rawdownloadcloneembedreportprint, ause, sburg \\
12 & anne, \textbackslash u0639,  sprayed & \textbackslash u30fc\textbackslash ufffd, \textbackslash ufffd\textbackslash ufffd, iltration \\
13 &  ear,  audio, Ear &  ear,  ears, Ear \\
14 &  goggles,  watched,  devotion & ideos, TPS,  goggles \\
15 & leaf,  slashed,  hunger &  gou, ouri,  margins \\
16 &  voices,  voic,  Hand &  voic,  leash,  voiced \\
17 &  short,  Short, short &  shorten,  shortened, short \\
18 & tones,  tones, tone &  bells, tone,  marrow \\
19 &  drawn,  connected, ieties & wu, River,  Awakening \\
\hline
\end{tabular}
}
\end{table}